\documentclass[onecolumn]{IEEEtran}

\usepackage{graphicx}
\DeclareGraphicsExtensions{.eps,.pdf,.png,.jpg,.jpeg}

\usepackage{multirow}

\usepackage[cmex10]{amsmath}
\usepackage{cite}
\usepackage{hyperref}

\begin{document}

\title{Performance Analysis on Evolutionary Algorithms for the Minimum Label Spanning Tree Problem}

\author{Xinsheng~Lai, Yuren~Zhou, Jun~He and Jun~Zhang
\thanks{X. Lai and Y. Zhou are with the School of Computer Science and Engineering, South China University of Technology, Guangzhou {\rm 510006}, China (e-mail:  yrzhou@scut.edu.cn).  Corresponding author: Y. Zhou}
\thanks{X. Lai is also with the School of Mathematics and Computer Science,
Shangrao Normal University, Shangrao 334001, China.}
\thanks{J. He is with the Department of Computer Science,
Aberystwyth University, Aberystwyth, SY23 3DB, UK.}
\thanks{J. Zhang is with the Department of Computer Science, Sun Yat-Sen
University, Guangzhou 510275, China, with the Key Laboratory of Digital
Life, Ministry of Education, China, and also with the Key Laboratory of
Software Technology, Education Department of Guangdong Province, Chin}
}

\maketitle

\begin{abstract}
Some experimental investigations have shown that evolutionary algorithms (EAs) are efficient for the minimum label spanning tree (MLST) problem. However, we know little about that in theory. As one step towards this issue, we theoretically analyze the performances of the (1+1) EA, a simple version of EAs, and a multi-objective evolutionary algorithm called GSEMO on the MLST problem. We reveal that for the MLST$_{b}$ problem the (1+1) EA and GSEMO achieve a $\frac{b+1}{2}$-approximation ratio in expected polynomial times of $n$ the number of nodes and $k$ the number of labels. We also show that GSEMO achieves a $(2ln(n))$-approximation ratio for the MLST problem in expected polynomial time of $n$ and $k$. At the same time, we show that the (1+1) EA and GSEMO outperform local search algorithms on three instances of the MLST problem. We also construct an instance on which GSEMO outperforms the (1+1) EA. 
\end{abstract}

\begin{IEEEkeywords}
Evolutionary algorithm; time complexity; approximation ratio; minimum label spanning tree; multi-objective
\end{IEEEkeywords}

\section{Introduction}
The minimum label spanning tree (MLST) problem is an issue arising from practice, which seeks a spanning tree with the minimum number of labels in a connected undirected graph with labeled edges. For example, we want to find a spanning tree that uses the minimum number of types of communication channels in a communication networks connected with different types of channels. The MLST problem, proposed by Chang and Leu, is proved to be NP-hard \cite{Chang}.

For this problem, Chang and Leu have proposed two heuristic algorithms. One is the edge replacement algorithm, ERA for short, the other is the maximum vertex covering algorithm, MVCA for short. Their experimental results showed that ERA is not stable, and MVCA is more efficient.

The genetic algorithm, belonging to the larger class of EAs, is a general purpose optimization algorithm \cite{Holland,Goldberg,Herrera} with a strong globally searching capacity \cite{Gallagher}. So, Xiong, Golden, and Wasil proposed a one-parameter genetic algorithm for the MLST problem. The experimental results on extensive instances generated randomly showed that the genetic algorithm outperforms MVCA \cite{Xiong}. Nummela and Julstrom also proposed an efficient genetic algorithm for solving the MLST problem \cite{Nummela}.

Besides, many methods recently have been proposed for solving this NP-hard problem. Consoli et al. proposed a hybrid local search combining variable neighborhood search and simulated annealing \cite{Consoli1}. Chwatal and Raidl presented exact methods including branch-and-cut and branch-and-cut-and-price \cite{Chwatal}. Cerulli et al. utilized several metaheuristic methods for this problem, such as simulated annealing, reactive tabu search, the pilot method, and variable neighborhood search \cite{Cerulli}. Consoli et al. still proposed a greedy randomized adaptive search procedure and a variable neighborhood search for solving the MLST problem \cite{Consoli2}.

Since both ERA and MVCA are two original heuristic algorithms for the MLST problem, the worst performance analysis of these two algorithms, especially MVCA, is a hot research topic in recent years. Krumke and Wirth proved that MVCA has a logarithmic performance guarantee of $2ln(n) + 1$, where $n$ is the number of nodes in the input graph,  and presented an instance to show that  ERA might perform as badly as possible \cite{Krumke}. Wan, Chen, and Xu further proved that MVCA has a better performance guarantee of $ln(n-1)+1$ \cite{Wan}. Xiong, Golden, and Wasil proved another bound on the worst performance of MVCA for MLST$_b$ problems, i.e., $H_b=\sum_{i=1}^b \frac{1}{i}$, where the subscript $b$ denotes that each label appears at most $b$ times, and also called the maximum frequency of the labels \cite{Xiong2}.

The performance of MVCA on the MLST problem has been deeply investigated. However, there is still no theoretical analysis work on EAs' performance for the MLST problem.

In fact, the theoretical analysis of EAs' performance on fundamental optimization problems has received much attention from many researchers. During the past few years theoretical investigations about EAs focused on the runtime or(and) the probability of EAs for finding globally optimal solutions of fundamental optimization problems or their variants. These problems include plateaus of constant fitness \cite{Jansen}, linear function problems \cite{He, Droste, He2}, minimum cut problems \cite{Neumann0}, satisfiability problems \cite{Zhou}, minimum spanning tree problems \cite{Neumann}, Eulerian cycle problems \cite{Neumann2}, Euclidean traveling salesperson problems \cite{Sutton}, etc.

Nevertheless, since many fundamental optimization problems, including the MLST problem, are NP-hard, no polynomial-time algorithm can be expected to solve them unless $P=NP$. Fortunately, we usually only ask satisfying solutions to such NP-hard problems in practice. Thus, we are interested in whether an approximation solution with a given satisfying quality can be efficiently obtained. In fact, the approximation performance analysis of randomized heuristics, including EAs, on NP-hard problems receives many attentions.

Giel and Wegener proved that the (1+1) EA can find a $(1+\varepsilon)$-approximation solution in expected runtime $O(m^{2\lceil1/\varepsilon\rceil})$, and concluded that EAs are good approximation algorithms for the maximum matching problem \cite{Giel}.

Subsequently, Oliveto, He, and Yao found that for minimum vertex cover problems the (1+1) EA may find arbitrary bad approximation solutions on some instances, but can efficiently find the minimum cover of them by using a restart strategy \cite{Oliveto}.  Friedrich et al. proved that the (1+1) EA may find almost arbitrarily bad approximation solution for minimum vertex cover problems and minimum set cover problems as well \cite{Friedrich}. Witt proved that in the worst case the (1+1) EA and the randomized local search algorithm need an expected runtime $O(n^2)$ to produce a $\frac{4}{3}$-approximation solution to the partition problem \cite{Witt}.

On the approximation performance of multi-objective EAs, Friedrich et al. revealed that the multi-objective EA efficiently finds an $(ln(n))$-approximation solution to the minimum set cover problem.  Neumann and Reichel found that multi-objective EAs can find a $k$-approximation solution for the minimum multicuts problem in expected polynomial time \cite{Neumann3}. Recently, Yu, Yao, and Zhou studied the approximation performance of SEIP, a simple evolutionary algorithm with isolated population, on set cover problems. They found that SEIP can efficiently obtain an $H_n$-approximation solution for unbounded set cover problems, and an $(H_n-\frac{k-1}{8k^9})$-approximation solution for $k$-set cover problems as well \cite{Yu2}.

In this paper, we concentrate on the performance analysis of the (1+1) EA and GSEMO for the MLST problem. We analyze the approximation performances of the (1+1) EA and GSEMO on the MLST problem. For the MLST$_b$ problem, We prove that the (1+1) EA and GSEMO are $\frac{b+1}{2}$-approximation algorithms. We also reveal that GSEMO can efficiently achieve a $(2ln(n))$-approximation ratio for the MLST problem. Though the MLST problem is NP-hard, we show that on three instances the (1+1) EA and GSEMO efficiently finds the global optima, while local search algorithms may be trapped in local optima. Meanwhile, we construct an additional instance where GSEMO outperforms the (1+1) EA.

The rest of this paper is organised as follows. The next section describes the MLST problem, and the algorithms considered in this paper. Section \uppercase\expandafter{\romannumeral3} analyzes the approximation performances of the (1+1) EA and GSEMO on the MLST problem, while section \uppercase\expandafter{\romannumeral4} analyzes the performances of the (1+1) EA and GSEMO on four instances. Finally, the section \uppercase\expandafter{\romannumeral5} presents the conclusions.

\section{The MLST problem and algorithms}

First of all, we give the concept of spanning subgraph.

\newtheorem{definition}{Definition}
\begin{definition}{(\textbf{Spanning subgraph})}
Let $G=(V,E)$ and $H=(V',E')$ be two graphs, where $V$ and $V'$ are, respectively, the sets of nodes of $G$ and $G'$, $E$ and $E'$ are, respectively, the sets of edges of $G$ and $G'$, if $V'=V$ and $E'\subset E$, then $H$ is a spanning subgraph of $G$.
\end{definition}

Let $G=(V,E,L)$ be a connected undirected graph, where $V$, $E$, and $L=\{1,2,\dots,k\}$ are the set of nodes, the set of edges, and the set of labels, respectively, $|V|=n$, $|E|=m$, and clearly $|L|=k$, each edge associates with a label by a function $l: E\rightarrow N$. Thus,
each edge $e\in E$ has an unique label $l(e)\in L$. The MLST problem is to seek a spanning tree with the minimum number of labels in the input graph $G$. If the maximum frequency of the labels is $b$, then we denote such an MLST problem by MLST$_{b}$. Clearly, the MLST$_b$ problem is a special case of the MLST problem.

Our goal in this paper is to seek a connected spanning subgraph with the minimum number of labels rather than a spanning tree with the minimum number of labels, since any spanning tree contained in such a spanning subgraph is a MLST. This is an alternative formulation of the MLST problem which is also adopted in papers \cite{Xiong, Nummela}.

We encode a solution as a bit string $X=(x_1,\dots,x_k)(\in \{0,1\}^k)$ which is used in \cite{Xiong}, where bit $x_i (1\leq i \leq k)$ corresponds to label $i$. If $x_i=1 (i=1,2,\dots,k)$, then label $i$ is selected, otherwise it is not. Thus, a bit string $X$ represents a label subset, and $|X|$ represents the number of labels contained in $X$.

We consider the spanning subgraph $H(X)$ of $G$, where $H(X)$ is a spanning subgraph restricted to edges with labels that the corresponding bits in $X$ are set to 1. We call a solution $X$ such that $H(X)$ is a connected spanning subgraph a feasible solution. A feasible solution with the minimum number of labels is a globally optimal solution.

For solving the MLST problem, the (1+1) EA  uses a fitness function, which is defined as
\begin{equation}
fit(X)=(c(H(X))-1)*k^2 + |X|,
\label{fitness1}
\end{equation}
where $c(H(X))$ is the number of connected components in $H(X)$, $k$ is the total number of labels in $L$, and $|X|=\sum_{i=1}^k x_i$, i.e, the number of labels conained in $X$ and also used in $H(X)$.

The fitness function should be minimized. The first part of it is to make sure that $H(X)$ is a connected spanning subgraph, and the second part is to make sure that the number of labels in the connected spanning subgraph is minimized.

For a feasible solution $X$, since the number of connected components of $H(X)$ is 1, the fitness value equals to the number of labels contained in it.

We also define the fitness vector for GSEMO as a vector $(c(H(X), |X|)$, where $c(H(X))$ and $|X|$ are simultaneously minimized by GSEMO.

The following algorithms are those considered in this paper.

\noindent \textbf{Algorithm 1: The (1+1) EA for the MLST problem}\\
01: \textbf{\emph{Begin}} \\
%\indent Initialize a solution $X=(x_1,x_2,\dots,x_k)$ uniformly at random;\\
02: \indent Initialize a solution $X\in\{0,1\}^k$ uniformly at random;\\
03: \indent \textbf{\emph{While}} termination criterion is not fulfilled\\
04: \indent \indent Obtain an offspring $Y$ by flipping each bit in $X$ with\\
      \indent \indent \indent \indent probability $\frac{1}{k}$;\\
05: \indent \indent \textbf{\emph{If}} $fit(Y) < fit(X)$ \textbf{\emph{then}} $X:=Y$;\\
06: \indent \textbf{\emph{End while}}\\
07: \textbf{\emph{End}}

The (1+1) EA starts with an arbitrary solution, and repeatedly uses mutation operator to generate an offspring solution from the current one. If the offspring solution is strictly better than the current one, then the (1+1) EA uses it to replace the current solution.

Another algorithm proposed by Br\"{u}ggemanna, Monnot, and Woeginger is called the local search algorithm with the $2$--switch neighborhood. We now describe some concepts about it.

\begin{definition}\cite{Bruggemann}{(\textbf{$h$-switch neighborhood})}
Given an integer $h\geq 1$, let $X_1$ and $X_2$ be two feasible solutions for some instance of the MLST problem. We say that $X_2$ is in $h$-switch neighborhood of $X_1$, denoted by $X_2\in h$-$SWITCH(X_1)$, if and only if
\begin{equation}
|X_1 - X_2|\leq h ~~and~~ |X_2-X_1|\leq h.
\end{equation}
\end{definition}

In other words, $X_2\in h$-$SWITCH(X_1)$ means that $X_2$ can be derived from $X_1$ by first removing at most $h$ labels from $X_1$ and then adding at most $h$ labels to it.

\textbf{The local search algorithm with the $2$-switch neighborhood}:

In the algorithm 1, if the initial solution $X$ is an arbitrary feasible solution, and the offspring $Y$ is selected from the $2$-switch neighborhood of $X$, then it is the local search algorithm with the $2$-switch neighborhood \cite{Bruggemann}.

GSEMO has been investigated on covering problems \cite{Friedrich}, pseudo-Boolean functions \cite{Giel2,Laumanns}, and minimum spanning tree problems \cite{Neumann4,Neumann5}. It is described as follows.

\noindent \textbf{Algorithm 2: GSEMO for the MLST problem}\\
01: \textbf{\emph{Begin}} \\
02: \indent Initialize a solution $X\in \{0,1\}^k$ uniformly at random;\\
03: \indent $P\leftarrow \{X\}$;\\
04: \indent \textbf{\emph{While}} termination criterion is not fulfilled\\
05: \indent \indent Choose a solution $X$ from $P$ uniformly at random;\\
06: \indent \indent Obtain an offspring $Y$ by flipping each bit in $X$ with\\
      \indent \indent \indent \indent probability $\frac{1}{k}$;\\
07: \indent \indent \textbf{\emph{If}} $Y$ is not dominated by $\forall X \in P$ \textbf{\emph{then}} \\ %any other solution in $P$ \\$X$ \textbf{\emph{then}} \\%by\\
08: \indent \indent \indent $Q:=\{X|X\in P,$ and $Y$ dominates $X$ $\}$;\\
09: \indent \indent \indent $P\leftarrow P\cup \{Y\}\setminus Q $;\\
10: \indent \indent \textbf{\emph{End if}}\\
11: \indent \textbf{\emph{End while}}\\
12: \textbf{\emph{End}}

In algorithm 2, $P$ is a population used to preserve those solutions which can not be dominated by any other from the population. The concept of domination is defined as follows.

Suppose the fitness vectors of solutions $X$ and $Y$ are $(c(H(X)),|X|)$ and $(c(H(Y)),|Y|)$, respectively. We say that $X$ dominates $Y$, if one of the following two conditions is satisfied:

\noindent (1) $c(H(X))<c(H(Y))$ and $|X|\leq |Y|$;\\
(2) $c(H(X))\leq c(H(Y))$ and $|X|<|Y|$.

For the sake of completeness, another two greedy algorithms are included.

The first one is \textbf{the modified MVCA}. It starts with a solution containing no labels, and each time selects a label such that when this label is chosen the decrease in the number of connected components is the largest.

\noindent \textbf{Algorithm 3: The modified MVCA for the MLST problem} \cite{Xiong2}\\
\noindent \textbf{\emph{Input:}} A given connected undirected graph $G=(V,E,L)$,\\
         \indent \indent \indent $L=\{1,\dots,k\}$.\\
01: \indent Let $C$ be the set of used labels, $C:=\emptyset$;\\
02: \indent \textbf{\emph{Repeat}}\\
03: \indent \indent Let $H$ be the spanning subgraph of $G$ restricted to\\
        \indent \indent \indent \indent edges with labels from $C$;\\
04: \indent \indent \textbf{\emph{For}} all $i\in L\setminus C$ \textbf{\emph{do}}\\
05: \indent \indent \indent Determine the number of connected components\\
        \indent \indent \indent \indent \indent when inserting all edges labeled by $i$ in $H$;\\
06: \indent \indent \textbf{\emph{End for}}\\
07: \indent \indent Choose label $i$ with the smallest resulting number of\\
        \indent \indent \indent \indent connected components: $C:= C\cup\{i\}$;\\
08: \indent \textbf{\emph{Until}} $H$ is connected.\\
\textbf{\emph{Output:}} $H$

In algorithm 3, if we contract each connected component in $H$ to a supernode after step 3, then we obtain the second greedy algorithm which is investigated in \cite{Krumke}, and we call it \textbf{the modified MVCA with contraction} in this paper.

\section{Approximation performances of the (1+1) EA and GSEMO on the MLST problem}

The following is the concept of approximation ratio (solution). Given a minimization problem $\textbf{P}$ and an algorithm $A$, if for an instance $\textbf{I}$ of $\textbf{P}$, the value of the best solution obtained in polynomial time by $A$ is $A(\textbf{I})$, and $sup_{\textbf{I}\in \textbf{P}}\frac{A(\textbf{I})}{OPT(\textbf{I})}=r$, where $OPT(\textbf{I})$ is the value of the optimal solution of $\textbf{I}$, then we say that $A$ achieves an $r$-approximation ratio (solution) for $\textbf{P}$.%, and that $A$ is a $r$-approximation algorithm for problem $\textbf{P}$.

Although the MLST problem is NP-hard, we reveal that the (1+1) EA and GSEMO guarantee to achieve an approximation ratio for the MLST$_b$ problem in expected polynomial times of $n$ and $k$, and that GSEMO guarantees to obtain an approximation ratio for the MLST problem in expected polynomial time of $n$ and $k$.

\subsection{The approximation guarantees of the (1+1) EA and GSEMO on the MLST$_b$ problem}%

To reveal that the (1+1) EA and GSEMO guarantee to achieve a $\frac{b+1}{2}$-approximation ratio in expected polynomial time of $n$ the number of nodes and $k$ the number of labels, we first prove that the (1+1) EA and GSEMO find a feasible solution starting from any initial solution in expected polynomial time of $n$ and $k$, then prove that starting from any feasible solution the (1+1) EA and GSEMO find a $\frac{b+1}{2}$-approximation solution in expected polynomial time of $k$ by simulating the following result proved by Br\"{u}ggemann, Monnot, and Woeginger \cite{Bruggemann}.

\newtheorem{theorem}{Theorem}
\begin{theorem}
If $b\geq 2$, then for any instance of the MLST$_{b}$ problem, the local search algorithm with the $2$-switch neighborhood can find a local optimum with at most $OPT\cdot \frac{b+1}{2}$ labels, where $OPT$ is the number of labels in the global optimum.
\label{thrmBruggemann}
\end{theorem}

We partition all feasible solutions into two disjoint sets. One is $S_1=\{X|X\in \{0,1\}^k$, $X$ is a feasible solution, $|X|\leq OPT\cdot \frac{b+1}{2}\}$, the other is $S_2=\{X|X\in \{0,1\}^k$, $X$ is a feasible solution, $|X| > OPT\cdot \frac{b+1}{2}\}$.

From theorem \ref{thrmBruggemann}, we derive a property with respect to the $2$-switch neighborhood for MLST$_{b}$ problems.

\newtheorem{corollary}{Corollary}
\begin{corollary}
If $b\geq 2$, let $G=(V,E,L)$ be an instance of MLST$_{b}$, which has a minimum label spanning tree with $OPT$ labels. If $X$ is a feasible solution, and $X\in S_2$, then there must exist a feasible solution $X'\in 2$-$SWITCH(X)$ whose fitness is 1 or 2 less than that of $X$.
\label{colFromthrmBruggemann}
\end{corollary}

We now prove that starting with an arbitrary initial solution for an instance $G=(V,E,L)$ of MLST$_b$, the (1+1) EA can efficiently find a feasible solution.

\newtheorem{lemma}{Lemma}
\begin{lemma}
Given an instance $G=(V,E,L)$ of MLST$_b$, where $|V|=n$, and $|L|=k$, the (1+1) EA starting from an arbitrary initial solution finds a feasible solution in $O(nk)$ for $G$.
\label{EAConnectedSpaningSubgraph}
\end{lemma}

\begin{IEEEproof}
According to the fitness function (\ref{fitness1}), during the optimization process of the (1+1) EA,  the number of connected components will never be increased.

For any solution $X$, if it is not a feasible solution, then the number of connected components of the spanning subgraph $H(X)$ is greater than 1. Since the input graph is connected, there must exist a label such that when it is added to the current solution $X$ the number of connected components will be decreased by at least one. The probability of adding this label to $X$ is $\frac{1}{k}(1-\frac{1}{k})^{k-1}\geq \frac{1}{ek}$, which implies that in expected time $O(k)$ the number of connected components will be decreased by at least one.

Note that there are at most $n$ connected components. Hence, we obtain the upper bound of $O(nk)$.
\end{IEEEproof}

Then, we prove that starting with an arbitrary feasible solution on an instance $G=(V,E,L)$ of MLST$_b$, the (1+1) EA can efficiently find a $\frac{b+1}{2}$-approximation solution.

\begin{lemma}
Given an instance $G=(V,E,L)$ of MLST$_b$, where $|V|=n$, $|L|=k$, and $b\geq 2$, the expected time for the (1+1) EA starting from an arbitrary feasible solution to find a local optimum with at most $OPT\cdot \frac{b+1}{2}$ labels for $G$ is $O(k^4)$.
\label{EAFindApprSlut}
\end{lemma}

\begin{IEEEproof}
By corollary \ref{colFromthrmBruggemann}, if a feasible solution $X\in S_2$, there must exist a feasible solution $X'\in 2$-$SWITCH(X)$ whose fitness is 1 or 2 less than that of $X$. So, replacing $X$ with $X'$ decreases the fitness value by at least 1. Since a feasible solution belonging to $S_2$ has at most $k$ labels, then after at most $k-OPT\cdot \frac{b+1}{2}$ such replacing steps a feasible solution belonging to $S_1$ will be found.

Now, we calculate the expected time for the (1+1) EA to find $X'$. Since $X'\in 2$-$SWITCH(X)$ and $|X'|<|X|$. There exist three cases. The first is that $X'$ is obtained by removing one exact label from $X$. The second is that $X'$ is obtained by removing two exact labels from $X$. The third is that $X'$ is obtained by removing two exact labels from $X$ and adding one exact label to it.

Obviously, the worst case is the third one, since in this case three bits of $X$ must  be simultaneously flipped by the (1+1) EA. In this case, the probability that for the (1+1) EA to find $X'$ is $\frac{1}{k^3}(1-\frac{1}{k})^{k-3}\geq \frac{1}{ek^3}$. So, the expected time for the (1+1) EA to find a feasible solution $X'\in 2$-$SWITCH(X)$ is $O(k^3)$, which means that the expected time for the (1+1) EA to reduce the fitness value by at least one is $O(k^3)$.

Therefore, the expected time for the (1+1) EA starting from an arbitrary feasible solution to find a local optimum with at most $OPT\cdot \frac{b+1}{2}$ labels is $O((k-OPT\cdot \frac{b+1}{2})k^3)=O(k^4)$, as $OPT\cdot \frac{b+1}{2}\leq k$.
\end{IEEEproof}

Combining Lemma \ref{EAConnectedSpaningSubgraph} and \ref{EAFindApprSlut}, we obtain the following Theorem.

\begin{theorem}
Given an instance $G=(V,E,L)$ of MLST$_b$, where $|V|=n$, $|L|=k$, and $b\geq 2$, the (1+1) EA starting with any initial solution finds a $\frac{b+1}{2}$-approximation solution in expected time $O((n+k^3)k)$.
\end{theorem}

It has been proved that the (1+1) EA efficiently achieves a $\frac{b+1}{2}$-approximation ratio for the MLST$_b$ problem. As we will see below, GSEMO can also efficiently achieve this approximation ratio.

\begin{theorem}
Given an instance $G=(V,E,L)$ of MLST$_b$, where $|V|=n$, $|L|=k$, and $b\geq 2$, GSEMO starting with any initial solution finds a $\frac{b+1}{2}$-approximation solution in expected time $O(nk^2+k^5)$.
\end{theorem}

\begin{IEEEproof}
Starting with an arbitrary solution $X$ with fitness vector $(c(H(X)),|X|)$, if $c(H(X))>1$, then there exists a label $l$ such that when it is added the number of connected components will be reduced by at lease one, as the input graph is connected. The probability of selecting $X$ to mutate is $\Omega(\frac{1}{k})$, as the population size is $O(k)$, and the probability of flipping the bit corresponding to label $l$ is $\frac{1}{k}(1-\frac{1}{k})^{k-1}=\Omega(\frac{1}{k})$, so a solution $X'$ with fitness vector $(c(H(X')),|X|+1)$ will be included in expected time $O(k^2)$, where $c(H(X'))<c(H(X))$.

Since there is at most $n$ connected components in the spanning subgraph induced by any solution, a feasible solution will be included in expected time $O(nk^2)$.

Now a feasible solution $X$ is included in the population, if the number of labels contained in the feasible solution is greater than $\frac{b+1}{2}\cdot OPT$, then according to Corollary \ref{colFromthrmBruggemann} there exists a feasible solution $X'\in 2$-$SWITCH(X)$ such that $|X'|$ is at least 1 less than $|X|$. If such a solution $X'$ is found, it will replace $X$. According to the proof of Lemma $\ref{EAFindApprSlut}$, the expected time to find such a solution $X'$ is $O(k^3)$. Combining the expected time to select such a solution $X$ is $O(k)$, such a solution $X'$ will be included in expected time $O(k^4)$.

Therefore, a $\frac{b+1}{2}$-approximation solution will be included in expected time $O((k-\frac{b+1}{2}\cdot OPT)k^4)=O(k^5)$ once a feasible solution is found.

Hence, GSEMO starting with any initial solution will find a $\frac{b+1}{2}$-approximation solution in expected time $O(nk^2+k^5)$.
\end{IEEEproof}

\subsection{The approximation guarantee of GSEMO on the MLST problem}

Here we prove the approximation guarantee of GSEMO on the MLST problem by simulating the process of the modified MVCA with contraction.
Similar to Lemma 2 in \cite{Krumke}, we prove the following Lemma.

\begin{lemma}
Given a connected undirected graph $G=(V,E,L)$ having a minimum label spanning tree $T_{OPT}$ with $OPT$ labels, where $|V|=n$ and $n\geq 2$, there exists a label such that the number of connected components of the spanning subgraph restricted to edges with this label is not more than $\lfloor n(1-\frac{1}{2OPT})\rfloor$.
\label{Krumke}
\end{lemma}

\begin{IEEEproof}
In fact, the minimum label spanning tree $T_{OPT}$ of $G$ has exact $n-1$ edges, there must exist a label, say $j$, such that the number of edges in $T_{OPT}$ labeled by $j$ is at least $\lceil\frac{n-1}{OPT}\rceil$, so the number of connected components of the spanning subgraph, restricted to edges of label $j$, is not more than $n-\lceil\frac{n-1}{OPT}\rceil= \lfloor n(1-\frac{1}{OPT}) + \frac{1}{OPT}\rfloor$. When $n\geq 2$, we have $\lfloor n(1-\frac{1}{OPT}) + \frac{1}{OPT}\rfloor \leq \lfloor n(1-\frac{1}{2OPT})\rfloor$.
\end{IEEEproof}

Further, for a spanning subgraph $H(X)$ of $G=(V,E,L)$, we have the following Corollary.

\begin{corollary}
If $r$ the number of connected components of $H(X)$ is greater than $2$, then there is a label such that when it is added to $X$ the number of connected components will be reduced to not more than $\lfloor r(1-\frac{1}{2OPT})\rfloor$.
\label{KrumkeCol}
\end{corollary}

\begin{IEEEproof}
Contracting each connected component of $H(X)$ to a supernode, then $G$ is converted to $G'$ with $r$ nodes. Suppose the number of labels in the minimum label  spanning tree of $G'$ is $OPT'$, according to Lemma \ref{Krumke}, there is a label in $G'$ such that the number of connected components of the spanning subgraph, restricted to edges with this label, is not more than $\lfloor r(1-\frac{1}{2OPT'})\rfloor$. Noting that the number of labels of the minimum label spanning tree of $G$ is $OPT$, it is clear that $OPT'\leq OPT$. Thus, $\lfloor r(1-\frac{1}{2OPT'})\rfloor<$ $\lfloor r(1-\frac{1}{2OPT})\rfloor$. In other words, there is a label such that when it is added to $X$ the number of connected components of $H(X)$ will be reduced to not more than $\lfloor r(1-\frac{1}{2OPT})\rfloor$.
\end{IEEEproof}

Based on Corollary \ref{KrumkeCol}, we prove that the GSEMO guarantees to find a $(2ln(n))$-approximate solution in expected polynomial time of $n$ and $k$.

\begin{theorem}
Given an instance $G=(V,E,L)$ of MLST problems, where $|V|=n$ and $|L|=k$, the expected time that GSEMO starts with any initial solution to find a $(2ln(n))$-approximation solution for $G$ is $O(k^3 ln(n)+k^2ln(k))$.
\label{GSEMOpoly2}
\end{theorem}

\begin{IEEEproof}
We first reveal that GSEMO starting with any initial solution will find the all-zeros bit string in expected time $O(k^2ln(k))$, then reveal that GSEMO finds a $(2ln(n))$-approximation solution in expected time $O(k^3 ln(n))$ after the all-zeros bit string being included in the population. Combining them, we obtain the Theorem.

We now investigate the expected time that GSEMO starting from any initial solution finds the all-zeros bit string with Pareto optimal fitness vector $(n,0)$. Once it is found, it can never be removed from the population. If it is not included in the population, GSEMO can choose a solution $X$ from $P$ which contains the minimum number of labels among all solutions in the population with probability $\Omega(\frac{1}{k})$, as the population size is $O(k)$. The event of flipping one of $|X|$ bits whose value is 1 will decrease the number of labels, and the probability for this event is $\binom{|X|}{1}\frac{1}{k}(1-\frac{1}{k})^{k-1}\geq \frac{|X|}{ek}$. So, the expected time that GSEMO includes a solution which contains $|X|-1$ labels is $O(\frac{k^2}{|X|})$. Following this way, the all-zeros bit string will be found in expected time $O(\sum_{i=|X|}^{1}\frac{k^2}{i})=O(k^2ln(|X|))=O(k^2ln(k))$.

Now that the all-zeros bit string with fitness vector $(n,0)$ is included in the population. According to corollary \ref{KrumkeCol}, there is a label such that when it is added to the
all-zeros bit string the number of connected components will be reduced to not more than $n(1-\frac{1}{2OPT})$. The probability of choosing this label is $\frac{1}{k}(1-\frac{1}{k})^{k-1}=\Omega(\frac{1}{k})$. Since the population size is $O(k)$, the probability of finding the all-zeros bit string
is $\Omega(\frac{1}{k})$. So a solution $X^1$ with fitness vector $(c_1,1)$, where $c_1\leq \lfloor n(1-\frac{1}{2OPT})\rfloor \leq n(1-\frac{1}{2OPT})$
can be included in the population in expected time $O(k^2)$.

If $c_1\geq 2$, then there is still a label such that when it is added to $X^1$ the number of
connected components will be reduced to not more than $n(1-\frac{1}{2OPT})^2$. So a solution $X^2$ with fitness vector $(c_2,2)$, where
$c_2\leq n(1-\frac{1}{2OPT})^2$ can be included in the population in expected time $O(k^2)$ after $X^1$ being included in the population.

Similarly, after solution $X^{h-1}$ with fitness vector $(c_{h-1},h-1)$, where $c_{h-1}\leq n(1-\frac{1}{2OPT})^{h-1}$, being included in the population, if $c_{h-1}\geq 2$, then a solution $X^h$ with fitness vector $(c_h,h)$, where $c_h\leq n(1-\frac{1}{2OPT})^h$, will
be included in the population in expected time $O(k^2)$.

Since when $h=2OPTln(n)$, $n(1-\frac{1}{2OPT})^h\leq 1$. So, a connected spanning subgraph with at most $2OPTln(n)$ labels will be finally included in the population in expected time $O(hk^2)=O(2OPTk^2ln(n))=O(k^3ln(n))$ after the all-zeros bit string being included in the population.
\end{IEEEproof}

\begin{table}[ht]
\tabcolsep=1.5pt \footnotesize \caption{Approximation performances of the (1+1) EA and GSEMO. '$r$', 'upper bound', and `---' refer to the approximation ratio, upper bound of the expected time, and unknown, respectively.}
\begin{center}
\begin{tabular}{c||c|c||c|c}
\hline
            & \multicolumn{2}{c}{The (1+1) EA} & \multicolumn{2}{c}{GSEMO}        \\
\hline
            & $r$            & upper bound     &  $r$           & upper bound    \\
\hline
\hline
MLST$_b$    & $\frac{b+1}{2}$ &$O(nk+k^4)$       &$\frac{b+1}{2}$       &$O(nk^2+k^5)$        \\
\hline
MLST        &---          & --- & $2ln(n)$        &$O(k^3ln(n)+k^2ln(k))$    \\
\hline
\end{tabular}
\end{center}
\label{CompareApprxmt}
\end{table}

Table \ref{CompareApprxmt} summarizes the approximation performances of the (1+1) EA and GSEMO for the minimum label spanning tree problem. For the MSLT$_b$ problem, the (1+1) EA and GSEMO can efficiently achieve a $\frac{b+1}{2}$-approximation ratio. However, the order of the expected time of GSEMO is higher than that of the (1+1) EA, the reason is that GSEMO has to select a promising solution to mutate in a population of size $O(k)$. For the MLST problem, GSEMO efficiently achieves a $2ln(n)$-approximation ratio, but the approximation performance of the (1+1) EA is unknown.

\section{Performances of the (1+1) EA and GSEMO on four instances}

In this section, we firstly present an instance where GSEMO outperforms the (1+1) EA, then we show that the (1+1) EA and GSEMO outperform local search algorithms on three instances of the MLST problem. 

\subsection{An instance where GSEMO outperforms the (1+1) EA}

At first, we construct an instance $G'=\{V,E,L\}$ to show that GSEMO is superior to the (1+1) EA, where $L=\{1,\dots,k\}$.

\begin{figure*}
\centering
\includegraphics[width=15cm]{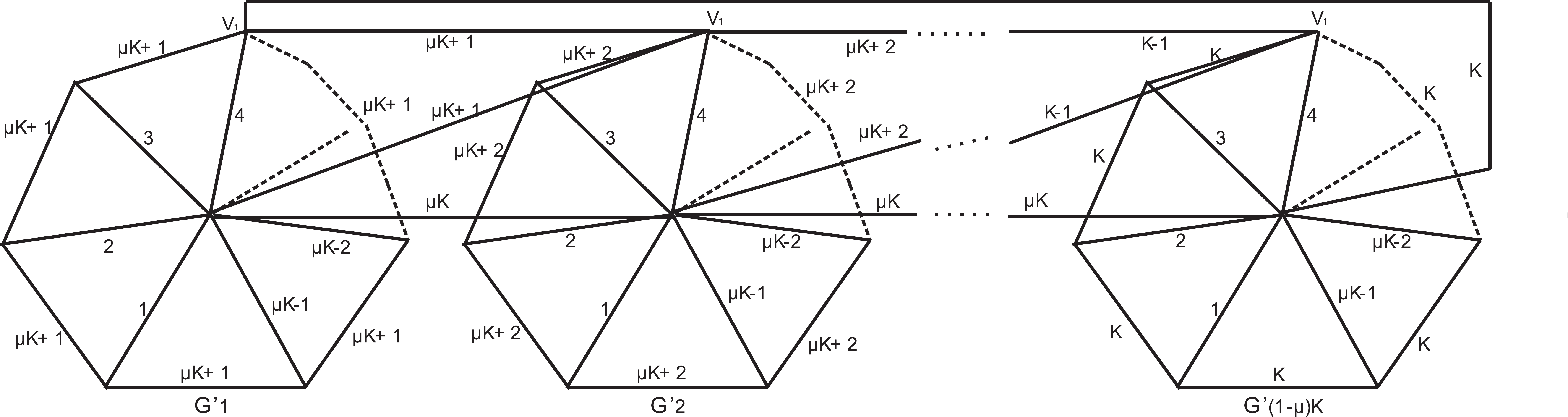}
\caption{An example of instance $G'$.}
\label{figEA}
\end{figure*}

Given $\mu(0<\mu<\frac{1}{2})$ and $k$ the number of labels, we construct instance $G'$ by the following steps. For simplicity, we assume that $\mu k$ is an integer, thus $(1-\mu)k$ is an integer as well. First, we construct $(1-\mu)k$ subgraphs $G'_1$, $\dots$, $G'_{(1-\mu)k}$. $G'_i$ $(1\leq i \leq (1-\mu)k)$ contains a $(\mu k-1)$-sided regular polygon whose edges are all labeled by the same label $\mu k+i$ and an inner node in the center. From the inner node, $(\mu k-1)$ edges labeled by from 1 to $\mu k-1$ connect to the $\mu k-1$ outer nodes $v_1$, $v_2$, $\dots$, $v_{\mu k-1}$. Then three edges are connected from $G'_i$ $(1\leq i\leq (1-\mu)k-1)$ to $G'_{i+1}$: the first one labeled by $\mu k+i$ is from the inner node of $G'_i$ to outer node $v_1$ of $G'_{i+1}$, the second one labeled by $\mu k+i$ is from outer node $v_1$ of $G'_i$ to outer node $v_1$ of $G'_{i+1}$, the third one labeled by $\mu k$ is from the the inner node of $G'_i$ to the inner node of $G'_{i+1}$. Finally, an additional edge labeled by $k$ connects the inner node of $G'_{(1-\mu)k}$ with outer node $v_1$ of $G'_1$. Figure \ref{figEA} shows instance $G'$.

For $0<\mu <1/2$, the global optimum of $G'$ is $X^*=(\overbrace{1,\dots,1}^{\mu k},\overbrace{0,\dots,0}^{(1-\mu)k})$, and the local optimum is $X^l=(\overbrace{0,\dots,0}^{\mu k},\overbrace{1,\dots,1}^{(1-\mu)k})$ for local search algorithms, such as the local search algorithm with the $2$-switch neighborhood, since both spanning subgraphs $H(X^*)$ and $H(X^l)$ are connected, but $|X^*|=\mu k$, $|X^l|=(1-\mu)k$, and $\mu k < (1-\mu)k$. For instance $G'$, the expected time for the (1+1) EA to jump out of the local optimum is exponential.

\begin{theorem}
For instance $G'$, starting from the local optimum $X^l$, the expected time for the (1+1) EA to find the global optimum is ${\Omega(k^{\mu k})}$.
\label{thmepisulon}
\end{theorem}

\begin{IEEEproof}
For instance $G'$, when the current solution is the local optimum $X^l$, the (1+1) EA only accepts the event that adds all $\mu k$ labels from $\{1,\dots,\mu k\}$ and simultaneously removes more than $\mu k$ labels from $\{\mu k+1,\dots,k\}$. So, the probability of escaping from the local optimum is

\indent \indent \indent \indent $\sum_{i=1}^{k-2\mu k}\binom{k-\mu k}{ \mu k+i}(\frac{1}{k})^{2\mu k+i}(1-\frac{1}{k})^{k-2\mu k-i}$ \\
\indent \indent \indent \indent $= (\frac{1}{k})^{\mu k}\sum_{i=1}^{k-2\mu k}\binom{k-\mu k}{\mu k+i}(\frac{1}{k})^{\mu k+i}(1-\frac{1}{k})^{k-2\mu k-i}$\\
\indent \indent \indent \indent $< (\frac{1}{k})^{\mu k}$.

This is because $\sum_{i=1}^{k-2\mu k}\binom{k-\mu k}{\mu k+i}(\frac{1}{k})^{\mu k+i}(1-\frac{1}{k})^{k-2\mu k-i}$ \\
\indent \indent \indent \indent \indent \indent \indent ~$<\sum_{i=1}^{k-2\mu k}\binom{k-\mu k}{\mu k+i}(\frac{1}{k})^{\mu k+i}(1-\frac{1}{k})^{k-2\mu k-i}\\
\indent \indent \indent \indent \indent \indent \indent ~+ \sum_{i=-\mu k}^{0}\binom{k-\mu k}{\mu k+i}(\frac{1}{k})^{\mu k+i}(1-\frac{1}{k})^{k-2\mu k-i}$\\
\indent \indent \indent \indent \indent \indent \indent ~~$=\sum_{i=0}^{k-\mu k}\binom{k-\mu k}{i}(\frac{1}{k})^{i}(1-\frac{1}{k})^{k-\mu k-i}=1$.

Thus, starting from the local optimum, the expected time for the (1+1) EA to find the global optimum of $G'$ is $O(k^{\mu k})$.
\end{IEEEproof}

Though the (1+1) EA needs expected exponential time to jump out of the local optimum, GSEMO can efficiently find the global optimum for instance $G'$.

\begin{theorem}
For instance $G'$, GSEMO finds the global optimum in expected time $O(k^2ln(k))$.
\label{GSEMOpoly}
\end{theorem}

\begin{IEEEproof}
Adding a label from $L_1=\{1,\dots,\mu k\}$ to the all-zeros bit string can reduce the number of connected components by $(1-\mu)k$, while adding a label from $L_2=\{\mu k+1,\dots,k\}$ can reduce the number of connected components by $\mu k$. Note that $(1-\mu)k$ is larger then $\mu k$, so, the Pareto front contains $\mu k +1$ Pareto optimal solutions with fitness vectors $(n,0)$, $(n-(1-\mu)k,1)$, $\dots$, $(n- (1-\mu)jk, j)$, $\dots$, $(1, \mu k)$, respectively. It is clear that the population size is $O(k)$.

It has been proved in Theorem \ref{GSEMOpoly2} that the expected time for GSEMO starting with any initial solution to include the all-zeros bit string in the population is $O(k^2ln(k))$.

Now we calculate the expected time to produce the whole Pareto front after the all-zeros bit string is found. The worst case is from the all-zeros bit string to produce the whole Pareto front. Suppose now in the population, there is a Pareto optimal solution $X$ with fitness vector $(n-(1-\mu)jk,j)$, which has the maximum number of labels. Another Pareto optimal solution with fitness vector $(n-(1-\mu)(j+1)k, j+1)$ can be produced by adding a label from $L_1$ which is not in $X$. The probability of adding this label is $\binom{\mu k-j}{ 1}\frac{1}{k}(1-\frac{1}{k})^{k-1}\geq \frac{\mu k-j}{ek}$. This implies that the expected time is $O(\frac{ek^2}{\mu k-j})$, as the expected time to select $X$ is $O(k)$. So, considering the worst case of starting from $(n,0)$, the expected time for GSEMO to produce the whole Pareto front is $\sum_{j=0}^{\mu k-1} \frac{ek^2}{\mu k-j}$$=O(k^2ln(k))$.
\end{IEEEproof}

\subsection{An instance where the (1+1) EA and GSEMO outperform ERA}

ERA is a local search algorithm. It takes an arbitrary spanning tree as input, then considers each non-tree edge and tests whether the number of used labels can be reduced by adding this non-tree edge and deleting a tree edge on the induced cycle.

\begin{figure}[ht]
\begin{center}
 \includegraphics[width=40mm]{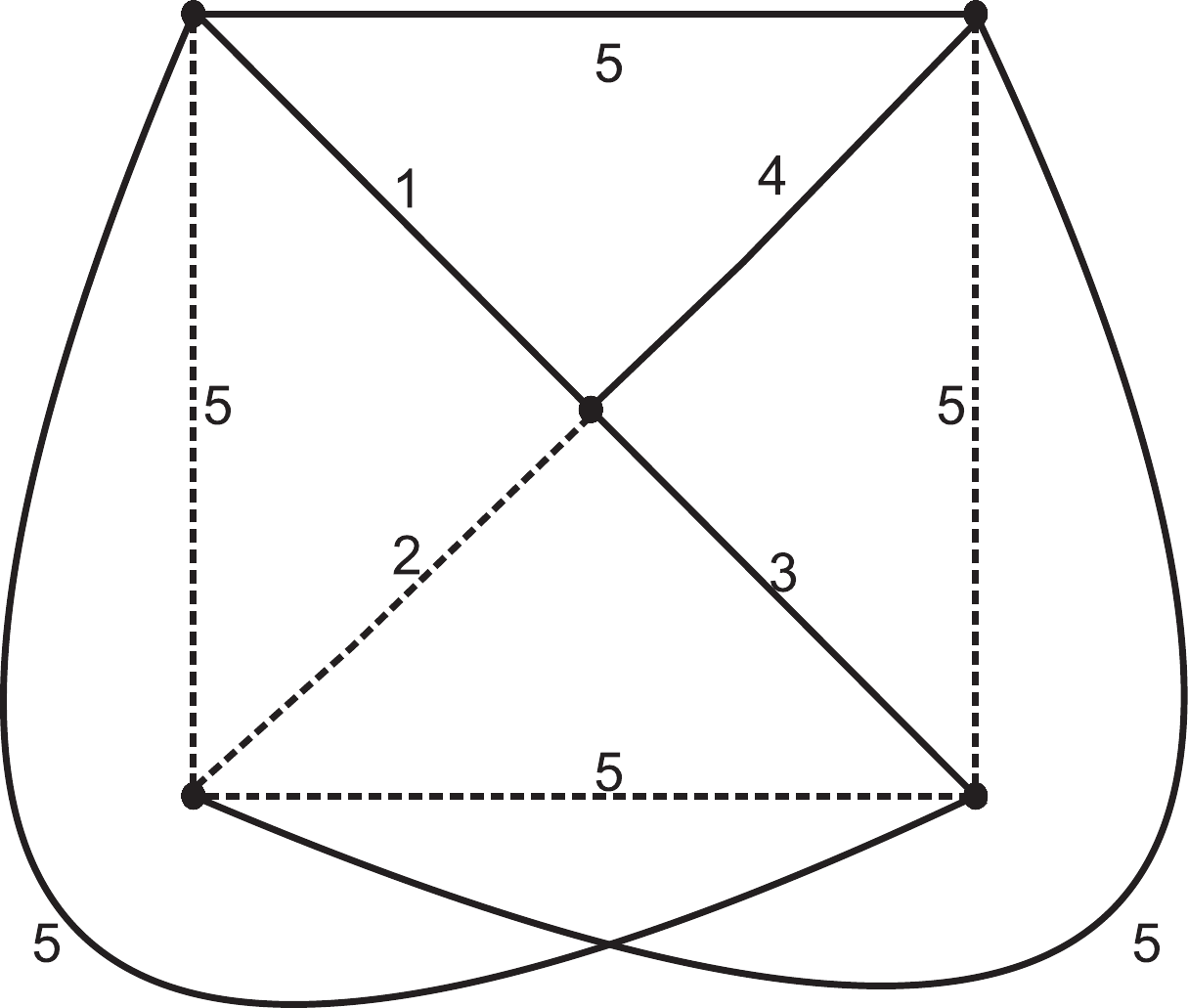}
\caption{An example of instance $G_1$ with $n=5$.}
\end{center}
\label{instance1}
\end{figure}

In this subsection, we show that the (1+1) EA and GSEMO outperform ERA on an instance proposed by Krumke and Wirth, which is denoted by $G_1$ in this paper.

This instance can be constructed by two steps. First, we construct a star shaped graph with $n-1$ distinct labels, i.e., selecting one node out of $n$ nodes, and adding $n-1$ edges from it to the other $n-1$ nodes labeled by $n-1$ distinct labels: $1$, $2$, $\dots$, $n-1$. Second, by adding edges between each pair of nodes with the same label $k$. Thus, we get a complete graph $G_1=(V,E,L)$, where $|V|=n$, $|E|=n(n-1)/2$, and $L=\{1,2,\dots,k\}$ is the set labels. It is clear that $|L|=k$, and $k=n$. Figure \ref{instance1} shows an example with $n=5$, where the dashed edges construct the spanning tree with the minimum number of labels.

For instance $G_1$, a global optimum $X^*$ uses one a label from $\{1,\dots,k-1\}$ and label $k$, i.e., $|X^*|=2$, $\sum_{i=1}^{k-1} x_i=1$, and $x_k=1$.

Krumke and Wirth used instance $G_1$ to demonstrate that ERA might perform as badly as possible. In fact, $X^l=(\overbrace{1,\dots,1}^{k-1},0)$ is a local optimum for ERA, since the number of labels used in $H(X^l)$ can not be reduced by adding any non-tree edge and deleting a tree edge on the induced cycle. The local optimum uses $k-1$ labels, while the global optimum uses only $2$ labels. However, the (1+1) EA and GSEMO can efficiently find a global optimum for $G_1$.

\begin{theorem}
For instance $G_1$, the (1+1) EA finds a global optimum in expected time $O(kln(k))$.
\label{instance1Thm}
\end{theorem}

\begin{IEEEproof}
For simplicity, let $L_1$ denote the label set $\{1,\dots,k-1\}$.

Let $A=\{X|c(H(X)=1,x_k=1, 2\leq |X|\leq k-1\}$, i.e., a solution $X\in A$ contains label $k$ and at least one but at most $k-2$ labels from $L_1$.

To find a global optimum, a solution $X\in A$ should be found first. Once a solution $X\in A$ has been found, the global optimum can be found by removing all $|X|-2$ redundant labels from $L_1$. According to the Coupon Collector's theorem \cite{Mitzenmacher}, all redundant labels contained in $X$ will be removed in expected time $O(kln(k))$.

In order to analyze the expected time to find a solution $X\in A$, we further partition all solutions that do not belong to $A$ into five disjoint subsets $B$, $C$, $D$, $E$, $F$: \\
$B=\{X|c(H(X))=1$, $|X|=k$, and $x_k=1\}$; \\
$C=\{X|c(H(X))=1$, $|X|=k-1$, and $x_k=0\}$; \\
$D=\{X|c(H(X))>1$, $1\leq |X|\leq k-2$, and $x_k=0\}$; \\
$E=\{X|c(H(X))>1$, $|X|=1$, and $x_k=1\}$; \\
$F=\{X|c(H(X))>1$, $|X|=0\}$.

If $X\in B$, then $X$ will be transformed into $A$ by removing one label from $L_1$. The probability of this event is $\binom{k-1}{ 1}\frac{1}{k}(1-\frac{1}{k})^{k-1}=\Omega(1)$, which implies that the expected time is $O(1)$.

If $X\in C$, then $X$ will be transformed into $A$ by adding label $k$ and simultaneously removing two labels from $L_1$. The probability of this event is $\binom{k-1}{2}(\frac{1}{k})^3(1-\frac{1}{k})^{k-3}=\Omega(\frac{1}{k})$, which implies that the expected time is $O(k)$.

If $X\in D(E)$, then $X$ will be transformed into $A$ by adding label $k$ (one label from $L_1$). The probability of this event is $\frac{1}{k}(1-\frac{1}{k})^{k-1}=\Omega(\frac{1}{k})$, which implies that the expected time is $O(k)$.

If $X\in F$, then $X$ will be transformed into $A$ by simultaneously adding label $k$ and a label from $L_1$. The probability is $\binom{k-1}{ 1}(\frac{1}{k})^2(1-\frac{1}{k})^{k-2}=\Omega(\frac{1}{k})$, which implies that the expected time is $O(k)$.

So, any solution will be transformed into $A$ in expected time $O(k)$.

Combining the expected time to remove all redundant labels contained in a solution belonging to $A$, the expected time for the (1+1) EA to find a global optimum is $O(kln(k))$.
\end{IEEEproof}

\begin{theorem}
For instance $G_1$, GSEMO finds a global optimum in expected time $O(k^2ln(k))$.
\label{instance1Thm2}
\end{theorem}

\begin{IEEEproof}
Let $L_1$ denote the label set $\{1,\dots,k-1\}$. We treat the optimization process as two independent phases: the first phase lasts until a solution with fitness vector $(1,.)$ is included, the second phase ends when a global optimum is found.

To analyze the expected time of the first phase, we consider the solution $X$ with fitness vector $(c(H(X)),|X|)$ where $c(H(X))$ is the minimum among all solutions in the population. If $c(H(X))>1$, then there are three cases. The first one is that $X$ contains no label, the second is that $X$ contains label $k$ but no label from $L_1$, the third is that $X$ contains at least one but at most $k-2$ labels from $L_1$ and no label $k$.

For all three cases, a solution with fitness vector $(1,.)$ will be included in expected time $O(k^2)$, since the probability of selecting $X$ to mutate is $\Omega(\frac{1}{k})$, and the probability of transforming $X$ into a solution with fitness vector $(1,.)$ is $\Omega(\frac{1}{k})$.

Once a solution with fitness vector $(1,.)$ is included, we show that a global optimum will be found in expected time $O(k^2ln(k))$. To this end, we partition the second phase into two subphases: the first subphase lasts until a solution belonging to $A=\{X|x_k=1, 2\leq |X|\leq k-1\}$ is found, i.e, such a solution contains label $k$ and at least one but at most $k-2$ labels from $L_1$, the second subphase ends when a global optimum is found.

If a solution $X$ with fitness vector $(1,.)$ and $X\not \in A$, then there are two cases needed to be considered: the first is that $X$ contains label $k$ and all labels from $L_1$, the other is that $X$ contains all labels from $L_1$ but no label $k$.

For the first case, removing any one of labels from $L_1$ will transform $X$ into $A$. The probability of this event is $\binom{k-1}{1}\frac{1}{k}(1-\frac{1}{k})^{k-1}$ $=\Omega(1)$, which implies that the expected time is $O(1)$. For the second case, removing two labels from $L_1$ and simultaneously adding label $k$ will transform $X$ into $A$. The probability of this event is $\binom{k-1}{2}(\frac{1}{k})^3(1-\frac{1}{k})^{k-3}=\Omega(\frac{1}{k})$, which implies that the expected time is $O(k)$. Combining the probability of selecting $X$ to mutate $\Omega(\frac{1}{k})$, a solution belonging to $A$ will be included in expected time $O(k^2)$ after a solution with fitness vector $(1,.)$ being included.

Now a solution $X\in A$ is included, the global optimum will be found by removing all $|X|-2$ redundant labels from $L_1$. If such a label is removed from $X$, then it can not be added any more. According to the Coupon Collector's theorem \cite{Mitzenmacher}, all redundant labels will be removed in expected mutations $O(kln(k))$, and the probability of selecting a solution belonging to $A$ to mutate is $\Omega(\frac{1}{k})$, so a global optimum will be found in expected time $O(k^2ln(k))$.

Therefore, GSEMO finds the global optimum in expected time $O(k^2ln(k))$.
\end{IEEEproof}

\subsection{An instance where the (1+1) EA and GSEMO outperform the local search algorithm with
the 2-switch neighborhood}

Br\"{u}ggemann, Monnot, and Woeginger proposed an instance, denoted by $G_2$ in this paper, to show that there exists a local optimum with respect to the local search algorithm with the 2-switch neighborhood \cite{Bruggemann}.

\begin{figure}[ht]
\begin{center}
\centerline{
\includegraphics[width=50mm]{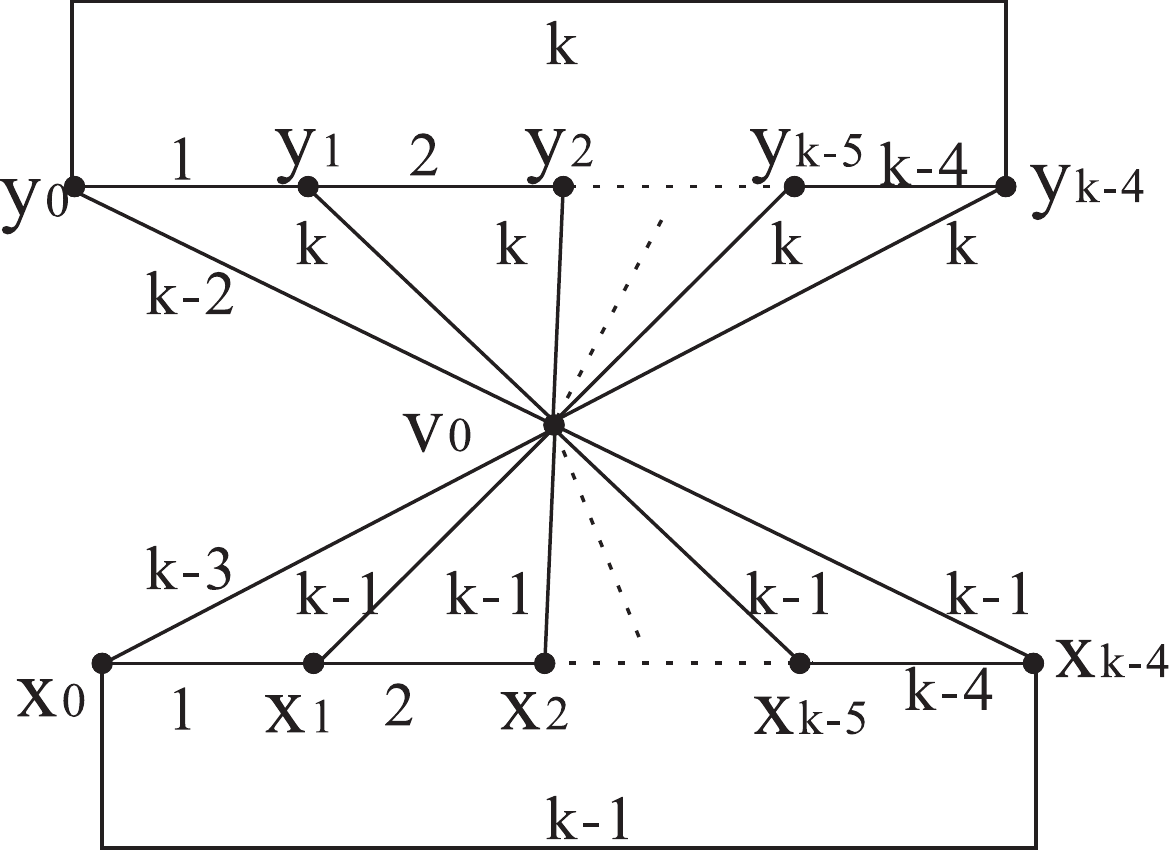}
}
 \caption{Instance $G_2$.} 
\label{instance2}
\end{center}
\end{figure}

\begin{figure}[h!]
\begin{center}
\centerline{
\includegraphics[width=50mm]{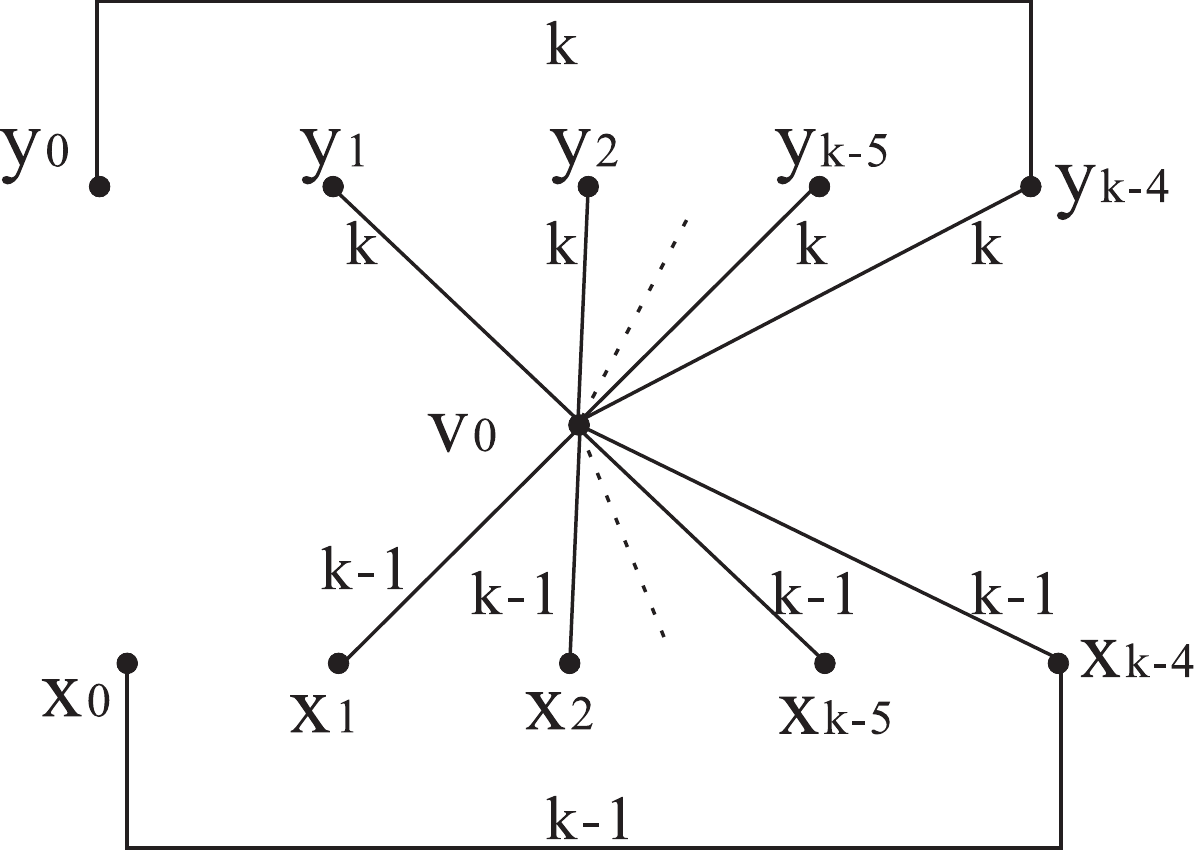}
}
\end{center}
\caption{The MLST of instance $G_2$.}
\label{instance2OPT}
\end{figure}

As shown in Figure \ref{instance2}, this instance is a graph $G_2=(V,E,L)$, where $V=($ $v_0$, $x_0$, $x_1$, $\dots$, $x_{k-4}$, $y_0$, $y_1$, $\dots$, $y_{k-4})$, $L=($$1$, $2$, $\dots$, $k)$, $|V|=2k-5$, $|E|=4k-12$,$|L|=k$. Figure \ref{instance2OPT} shows the minimum label spanning tree.

In this instance, the global optimum is $X^*=($$\overbrace{0,0,\dots,0}^{k-2}$, $1$, $1)$.

The local search algorithm with the $2$-switch neighborhood might be trapped in the local optimum which contains labels 1, 2, $\dots$, $k-2$. In fact, to jump out of this local optimum, at least three labels from $\{1,2,\dots,k-2\}$ should be removed and simultaneously two labels $k-1$ and $k$ should be added, but the resulting solution is not in the the $2$-switch neighborhood of the local optimum. However, the (1+1) EA and GSEMO can efficiently find the global optimum of $G_2$.

\begin{theorem}
For instance $G_2$, the (1+1) EA finds the global optimum in expected time $O(k^2)$.
\label{G21+1}
\end{theorem}

\begin{IEEEproof}
Let $L_1$ denote the label set $\{1,\dots,k-2\}$, and let $A=\{X|c(H(X))=1,x_{k-1}=1,x_k=1,2\leq |X|\leq k-1\}$, i.e, a solution $X\in A$ contains labels $k-1$ and $k$ and at most $k-3$ labels from $L_1$.

Noting that the global optimum contains only two labels $k-1$ and $k$, we treat the optimization process as two phases: the first phase lasts until a solution $X\in A$ is constructed from an arbitrary solution, the second phase ends when all $|X|-2$ redundant labels from $L_1$ are removed.

For analyzing the expected time of finding a solution $X\in A$, we partition all solutions that do not belong to $A$ into seven disjoint subsets $B$, $C$, $D$, $E$, $F$, $G$, $H$:

\begin{align*}
&B=\{X|c(H(X)=1, x_{k-1}=0,x_k=0, |X|=k-2\};\\
&C=\{X|c(H(X)=1, x_{k-1}=0,x_k=1, |X|=k-2 or |X|=k-1\};\\
&D=\{X|c(H(X)=1, x_{k-1}=1,x_k=0, |X|=k-2 or |X|=k-1\};\\
&E=\{X|c(H(X)=1, x_{k-1}=1,x_{k}=1, |X|=k\};\\
&F=\{X|c(H(X)>1, x_{k-1}=0,x_{k}=0\};\\
&G=\{X|c(H(X)>1, x_{k-1}=0,x_{k}=1\};\\
&H=\{X|c(H(X)>1, x_{k-1}=1,x_{k}=0\};
\end{align*}

If $X\in B$, then $X$ will be transformed into $A$ by adding labels $k-1$, $k$, and simultaneously removing three labels from $L_1$. The probability of this event is $\binom{k-2}{ 3} (\frac{1}{k})^5(1-\frac{1}{k})^{k-5}$$=\Omega(\frac{1}{k^2})$, which implies that the expected time is $O(k^2)$.

If $X\in C(D)$, then $X$ will be transformed into $A$ by adding label $k-1$ ($k$) and simultaneously removing two labels from $L_1$. The probability of this event is at least $\binom{k-3}{ 2}(\frac{1}{k})^3(1-\frac{1}{k})^{k-3}=\Omega(\frac{1}{k})$, which implies that the expected time is $O(k)$.

If $X\in E$, then $X$ will be transformed into $A$ by removing a label from $L_1$. The probability of this event is $\binom{k-2}{ 1}\frac{1}{k}(1-\frac{1}{k})^{k-1}=\Omega(1)$, which implies that the expected time is $O(1)$.

If $X\in F$, then $X$ will be transformed into $A$ by simultaneously adding labels $k-1$ and $k$. The probability of this event is $(\frac{1}{k})^2(1-\frac{1}{k})^{k-2}$$=\Omega(\frac{1}{k^2})$, which implies that the expected time is $O(k^2)$.

If $X\in G(H)$, then $X$ will be transformed into $A$ by adding label $k-1$ ($k$). The probability of this event is $\frac{1}{k}(1-\frac{1}{k})^{k-1}$$=\Omega(\frac{1}{k})$, which implies that the expected time is $O(k)$.

So, a solution belonging to $A$ will be found in expected time $O(k^2)$.

In the second phase, removing each label contained in a solution belonging to $A$ which is from $L_1$ will reduce the fitness value, and once it is removed it can not be added any more. According to the Coupon Collector¡¯s theorem \cite{Mitzenmacher}, the second stage ends in expected time $O(kln(k))$.

Altogether, the expected time for the (1+1) EA to find the global optimum is $O(k^2)$.
\end{IEEEproof}

\begin{theorem}
For instance $G_2$, the expected time for GSEMO to find the global optimum is $O(k^2ln(k))$.
\end{theorem}

\begin{IEEEproof}
It has been proved in Theorem \ref{GSEMOpoly2} that the expected time for GSEMO starting with any initial solution to find the all-zeros bit string is $O(k^2ln(k))$.

Once the all-zeros bit string is included in the population, the Pareto optimal solution $X^1$ with fitness vector $(k-2,1)$ will be found by adding label $k-1$ or $k$, to the all-zeros bit string. Since the probability of finding the all-zeros bit string is $\Omega(\frac{1}{k})$, and the probability of flipping a bit corresponding to such labels is $\frac{2}{k}(1-\frac{1}{k})^{k-1}=\Omega(\frac{1}{k})$. So, the expected time to produce $X^1$ from the all-zeros bit string is $O(k^2)$. Then the Pareto solution $X^2$ with fitness vector $(1,2)$ will be found by adding the remaining label from $\{k-1,k\}$ to solution $X^1$, and the expected time to produce solution $X^2$ from solution $X^1$ is also $O(k^2)$.

Therefore, the expected time for GSEMO to find the global optimum is $O(k^2ln(k))$.
\end{IEEEproof}

\subsection{An instance where the (1+1) EA and GSEMO outperform the modified MVCA}

\begin{figure*}
\centering
\includegraphics[width=130mm]{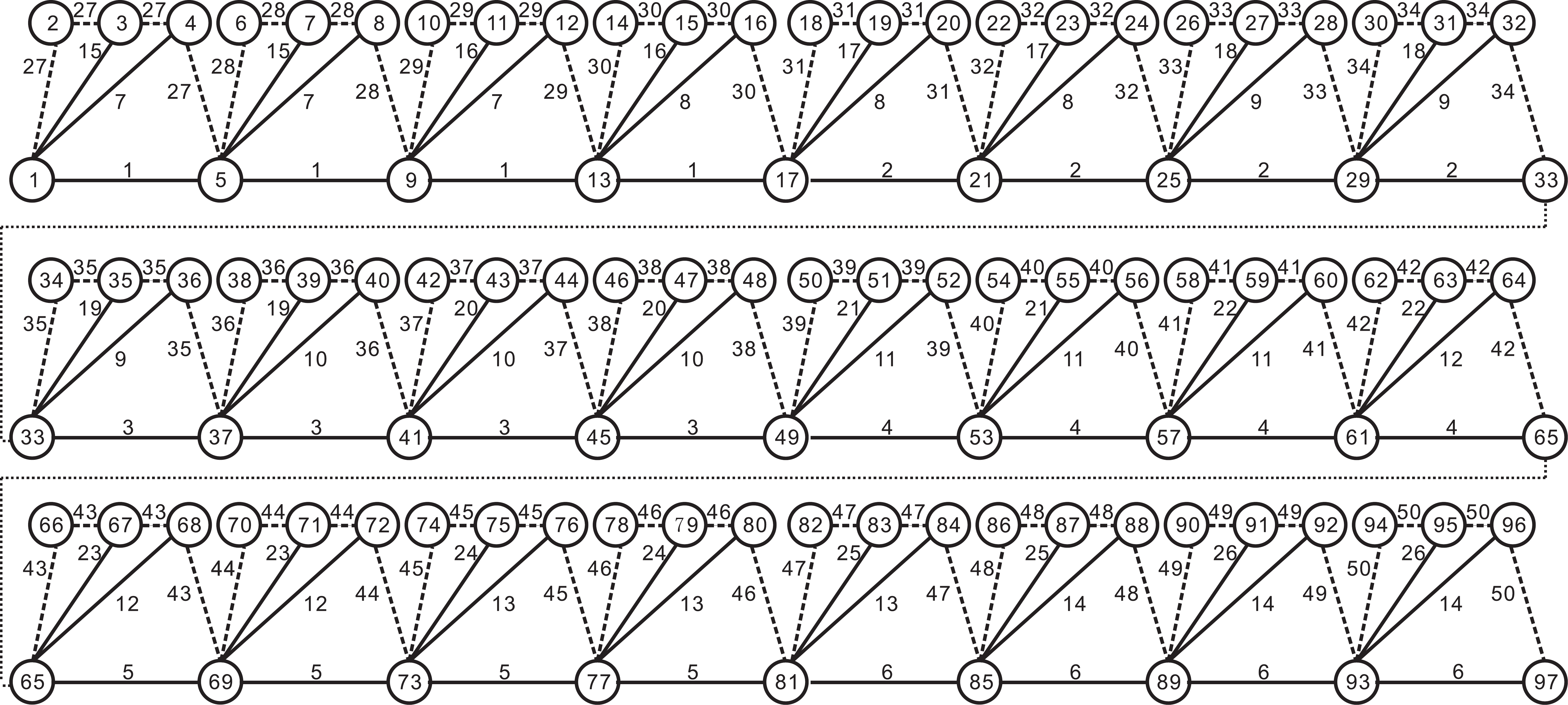}
\caption{Instance $G_3$ with b=4.}
\label{instance3}
\end{figure*}

In this subsection, we show that the (1+1) EA and GSEMO outperform the modified MVCA on an instance proposed by Xiong, Golden, and Wasil \cite{Xiong2}, which is denoted by $G_3$ in this paper.

Given the bound of the labels' frequency $b(b\geq 2)$, and let $n=b\cdot b!+1$, we construct $G_3=(V,E,L)$ as follows, where $V=\{1,2,\dots,n\}$, $|V|=n$, and $L=L_b\cup L_{b-1}\cup \dots \cup L_2\cup L_{opt}$.

We construct $b!$ groups from $V$, each containing $b+1$ nodes:

\noindent $V_1=\{1,2,\dots, b+1\}$,\\
$V_2=\{b+1,b+2,\dots, 2b+1\}$,\\
$\dots$\\
$V_j=\{(j-1)b+1,(j-1)b+2,\dots, jb+1\}$,\\
$\dots$\\
$V_{b!}=\{(b!-1)b+1, (b!-1)b+2, \dots, b!b+1\}$.

In $V_j$ $(j=1,2,\dots,b!)$, the edges between consecutive nodes $((j-1)b+1,(j-1)b+2)$, $\dots$, $(jb,jb+1)$ are all labeled with one label. Thus, $b!$ labels are needed, which constitute the label set $L_{opt}$. The edges with these $b!$ labels construct the minimum label spanning tree $T_{opt}$, so in this instance $OPT=b!$.

The label subset $L_h$ $(h=b, b-1,\dots, 2)$ is obtained as follows. We choose edge $((j-1)b+1,(j-1)b+1+h)$ in each $V_j$, so there are $b!$ such edges. We label the first $h$ edges with one label, and the next $h$ edges with a second label, etc. So, $\frac{b!}{h}$ labels are needed, and they construct $L_h$. Hence, $|L_h|=\frac{b!}{h}$, and the total number of labels $k=\sum_{j=2}^b \frac{b!}{j}+b!$.

Figure \ref{instance3} shows an example with $b=4$, where the dashed edges construct the spanning tree with the minimum number of labels.

In this instance, the global optimum is $X^*=(\overbrace{0,\dots,0}^{\sum_{j=2}^b\frac{b!}{j}},\overbrace{1,\dots,1}^{b!})$.

Xiong, Golden, and Wasil used this instance to show that the modified MVCA may obtain the worst-case solution using all labels from $L_b\cup L_{b-1}\cup\dots\cup L_2\cup L_{opt}$, which is $H_b$-approximation solution, where $H_b=\sum_{i=1}^b \frac{1}{i}$. Here, we show that the (1+1) EA and GSEMO can efficiently find the global optimum.

\begin{theorem}
For instance $G_3$, the (1+1) EA finds the global optimum in expected time $O(nk)$, where $n=b\cdot b!+1$, $k=\sum_{j=2}^b \frac{b!}{j} +b!$, and $b$ is the maximum frequency of the labels.
\label{emptyIni}
\end{theorem}

\begin{IEEEproof}
We treat the optimization process as two independent phases. The first phase ends when the (1+1) EA finds a solution $X$ such that $H(X)$ is a connected spanning subgraph. The second phase lasts until the (1+1) EA removes all redundant labels from $\{1,2, \dots, \sum_{j=2}^b \frac{b!}{j}\}$.

Let $X$ be the current solution, if the number of connected components of $H(X)$ is not $1$, then there must exist a bit $x_h$ from $\{x_i|\sum_{j=2}^b \frac{b!}{j}+1 \leq i \leq \sum_{j=2}^b \frac{b!}{j}+b!\}$ valued 0, otherwise the number of connected components is $1$. So, the (1+1) EA can decrease the number of connected components by at least $1$ with probability at least $\frac{1}{k}(1-\frac{1}{k})^{k-1}\geq \frac{1}{ek}$. This is the probability of the event that bit $x_h$ is flipped from $0$ to $1$ while the other bits keeping unchanged. Hence, the expected time to decrease the number of connected components from $n$ to $1$ is $O(nk)$, i.e., a connected spanning subgraph will be created by the (1+1) EA in expected runtime $O(nk)$.

Once a connected spanning subgraph is constructed, each bit from $\{x_i|\sum_{j=2}^b \frac{b!}{j}+1 \leq i \leq \sum_{j=2}^b \frac{b!}{j}+b!\}$ takes value 1, and the flippings of them can not be accepted by the (1+1) EA, as such flippings will create a disconnected spanning subgraph. For each bit $x_i (1\leq i \leq \sum_{j=2}^b \frac{b!}{j})$, if $x_i=1$, then it can be flipped from 1 to 0, since this will decrease the fitness value; otherwise, its flipping can not be accepted by the (1+1) EA, as this will increase the fitness value. So, when all bits have been selected at least once to flip, the connected spanning subgraph with the minimum number of labels will be found. According to the Coupon Collector's theorem \cite{Mitzenmacher}, the upper bound of the expected runtime for this to happen is $O(kln(k))$.

Hence, the expected time needed for the (1+1) EA to find the global optimum is $O(nk+kln(k))$ $=O(nk)$. Note that $n=b\cdot b!+1 > k=b!(1+\frac{1}{2}+\dots+\frac{1}{b})$, and $k> ln(k)$. So, $n> ln(k)$, and $nk>kln(k)$.
\end{IEEEproof}

GSEMO can also efficiently find the global optimum for $G_3$.

\begin{theorem}
For instance $G_3$, GSEMO finds the global optimum in expected time $O(k^3)$.
\end{theorem}

\begin{IEEEproof}
We treat the optimization process as two phases: the first phase is that GSEMO starting from any initial solution finds a solution with fitness vector $(1,.)$, the second phase is that GSEMO finds the global optimum after a solution with fitness vector $(1,.)$ being found.

Noting that a connected spanning subgraph contains all labels from $L_{opt}=\{\sum_{j=2}^b \frac{b!}{j}+1, \dots, \sum_{j=2}^b \frac{b!}{j}+b!\}$, if $X$ is a solution such that $c(H(X))>1$, then at least one labels from $L_{opt}$ are not contained in it, and the number of connected components can be decreased by adding such labels.

We now analyze the expected time that GSEMO starting from any initial solution finds a solution with fitness vector $(1,.)$. If such a solution has not been included in the population, then there is a solution $X$ from $P$ such that $c(H(X))$ is the minimal, and adding some label $l$ from $L_{opt}$ to $X$ will reduce the number of connected components. The probability that GSEMO chooses $X$ to mutate is $\Omega(\frac{1}{k})$, as the population size is $O(k)$, and the probability of flipping the bit corresponding to label $l$ is $\frac{1}{k}(1-\frac{1}{k})^{k-1}=\Omega(\frac{1}{k})$, so a solution with a smaller number of connected components will be found in expected time $O(k^2)$. After all labels from $L_{opt}$ being added, a connected spanning subgraph will be constructed. Thus a solution with fitness vector $(1,.)$ will included in expected time $O(b!k^2)=O(k^3)$.

Once a solution with fitness vector $(1,.)$ is included, GSEMO can finish the second phase by removing all redundant labels from $L_b\cup L_{b-1} \cup \dots \cup L_2$ one by one. Since the probability of selecting the solution with fitness vector $(1,.)$ to mutate is $\Omega(\frac{1}{k})$, and removing all redundant labels needs an expected time $O(kln(k))$. So the global optimum will be find in expected time $O(k^2ln(k))$.

Combining the expected times in two phases, we finish the proof.
\end{IEEEproof}

\begin{table}[ht]
\tabcolsep=1.5pt \footnotesize \caption{Upper bounds on the expected times for the (1+1) EA and GSEMO to find the global optima on four instances. `---' means unknown.}
\begin{center}
\begin{tabular}{c||c|c|c|c}
\hline
                & Instance $G'$ &Instance $G_1$ &Instance $G_2$ &Instance $G_3$        \\
\hline
\hline
The (1+1) EA    & ---           &$O(kln(k))$      &$O(k^2)$       &$O(nk)$        \\
\hline
GSEMO           & $O(k^2ln(k))$ &$O(k^2ln(k))$  &$O(k^2ln(k))$  &$O(k^3)$ \\
\hline
\end{tabular}
\end{center}
\label{Compare}
\end{table}

Table \ref{Compare} summarizes the upper bounds on the expected times for the (1+1) EA and GSEMO to find the global optima on all four instances. On instances $G_1$, $G_2$, and $G_3$, GSEMO needs times of higher order than the (1+1) EA. The main reason is that GSEMO selects a promising solution to mutate in a population of size $O(k)$. On $G'$, GSEMO outperforms the (1+1) EA because GSEMO behaves greedily and optimizes solutions with different number of labels.

\section{Conclusion}

In this paper, we investigated the performances of the (1+1) EA and GSEMO on the minimum label spanning tree problem. We found that the (1+1) EA and GSEMO can guarantee to achieve some approximation ratios. We further show that the (1+1) EA and GSEMO defeat local search algorithms on some instances, and that GSEMO outperforms the (1+1) EA on an instance.

As for the approximation ratio of the (1+1) EA on the MLST problem, we still know nothing about. Apart from this, since the (1+1) EA and GSEMO are randomized algorithms, it is natural to ask whether they can achieve better approximate ratios than those guaranteed by some greedy algorithms. From our analysis process, especially from the analysis process of GSEMO, it seems possible.

\paragraph*{Acknowledgement}
This work was supported in part by the National Natural Science Foundation
of China (61170081, 61165003, 61300044, 61332002), in part by the EPSRC
under Grant EP/I009809/1, in part by the National High-Technology Research
and Development Program (863 Program) of China No. 2013AA01A212,
and in part by the NSFC for Distinguished Young Scholars 61125205.

\end{document}